\documentclass{article}
\usepackage[preprint, nonatbib]{neurips_2019}

\usepackage{amsmath}
\usepackage{amssymb}
\usepackage{enumitem}
\usepackage{graphicx}
\usepackage{hyperref}
\usepackage{grffile}
\usepackage{subfigure}
\usepackage{comment}

\newcommand{\discriminator}{\mathcal{D}}
\newcommand{\priorKnowledgeSet}{T}
\newcommand{\originalSet}{X}

\title{Learning Interpretable Fair Representations}
\author{Tianhao Wang \thanks{indicates equal contribution} \And
Zana Bu\c cinca \footnotemark[1] \And
Zilin Ma \footnotemark[1]}
\date{May 2020}

\begin{document}

\maketitle

\begin{abstract}
Numerous approaches have been recently proposed for learning fair representations that mitigate unfair outcomes in prediction tasks. A key motivation for these methods is that the representations can be used by third parties with unknown objectives. 
However, because current fair representations are generally not interpretable, the third party cannot use the obtained representation for exploration, or to obtain any additional insight, besides the pre-contracted prediction tasks. Thus, to increase data utility beyond prediction tasks, we argue that the representations need to be fair, yet interpretable. We propose a general framework for learning interpretable fair representations by introducing an interpretable ``prior knowledge" during the representation learning process. 
We implement this idea and conduct experiments with ColorMNIST and Dsprite datasets. The results indicate that in addition to being intepretable, our representations attain slightly higher accuracy and fairer outcomes in a downstream classification task compared to state-of-the-art fair representations.
\end{abstract}

\section{Introduction}

Algorithmic fairness seeks to build fair classifiers that prevent any discrimination against subgroups of population varying on sensitive attributes such as race and gender. 
In advertising, prior work has considered classification as a process involving two distinct parties -- the data owner and the vendor \cite{dwork2012fairness}. The data owner profits by providing useful data to the vendor. Whereas, the vendor profits from advertising the appropriate products to consumers (i.e. prediction task). In these settings, the vendor might attempt to maximize their profits by disproportionally and unfairly showing distinct content to different groups. One way of avoiding unfair exploitation of the data is for the data owner to supply the vendor with fair representations that are also high in utility.

In this vein, numerous studies have proposed robust approaches for the data owner to transform its data into fair representations with little trade-off in vendor’s prediction accuracy \cite{madras2018learning,ruoss2020learning, lahoti2019operationalizing, feng2019learning, zemel2013learning}. However, while fair, these representations are generally not interpretable. Consequently, the vendor cannot interpret the data they are using for prediction, thus can only use the data for prediction tasks, and not for exploration. We argue that this might hinder the utility of the data, because the vendor cannot draw any additional insights from inspecting the representations. For example, different correlations can be found among attributes in fair and interpretable representations which might help the vendor refine their products. Therefore, we postulate that to increase the utility of the data beyond prediction tasks, we should aim for fair, yet interpretable representations.

In addition to extending the utility of the data for a vendor, interpretable fair representations can be invaluable when there is a human making decisions in tandem with the machine, as is the case for many of the high stakes decision-making scenarios such as recidivism prediction, loan approval and college application. Prior research has shown that in model assisted decision-making, where human is the final arbiter, human biases creep in even when the model is fair \cite{green2019disparate}. In these settings, human is presented with the original data and the model's decision. Note that there is no utility in showing fair, but not interpretable, representations to a decision-maker. Hence, if human is shown the interpretable fair representations along with model's decision, the overall decision of the sociotechnical system is expected be fair, because there is no way of inferring sensitive attributes. 


Taking the aforementioned motivations into account, we propose a simple yet effective approach to learn fair and interpretable representations by introducing ``prior knowledge". 
This ``prior knowledge" is a tailored representation that instantiates what the data owner \emph{thinks} a fair representation for the data should look like. 
For example, we experiment with a synthetic binary classification task with \texttt{ColorMNIST} dataset, where the color of the digit is considered as the sensitive attribute, whereas the shape of the digit is the one used for labeling. 
Here, the data owner would provide gray-scale images, which retain information only about the shapes of the digits, as ``prior knowledge" to the algorithm.
The final representations will seek to optimize fairness metrics (i.e. group fairness or individual fairness), while being constrained to resemble the given interpretable ``prior knowledge``. 
We also experiment with \texttt{DSprite} dataset, where we label the object by its shape but construct a (spurious) correlation between the label and object scale. 
We use demographic parity as the fairness metric in the experiment, but our framework can be easily adapted to individual fairness settings also. 
Our results indicate that our approach yields representations that attain higher accuracy, are fairer and much more interpretable than state-of-the-art baseline for fair representations.

\section{Background \& Related Work}

\textbf{Fairness Metrics.} A vital part of building fair algorithms is the selection of fairness metric. Broadly, fairness metrics are categorized as metrics that ensure \textit{group fairness} or \textit{individual fairness}. Group fairness metrics, such as \textit{demographic parity, equalized odds} and \textit{equal opportunity} are concerned with ensuring fairness in terms of the overall accuracy, number of false positive and number of false negatives amongst groups \cite{hardt2016equality}. Whereas, individual fairness ensures that similar individuals are treated similarly \cite{dwork2012fairness}. The downsides of group fairness metrics have been discussed widely \cite{dwork2012fairness, chouldechova2017fair, baer2019fairness}, and the challenges of coming up similarity metric in individual fairness have been highlighted \cite{dwork2012fairness}. 

\textbf{Fair Representations.} One way of warranting algorithmic fairness, defined either by group or individual fairness metrics, is by constructing fair data representations from which sensitive attributes cannot be derived. Because these representations need to maintain utility, while simultaneously satisfying fairness, numerous studies have utilized adversarial learning objectives to achieve the goal. The adversary's intent is to learn the sensitive attributes, whereas the fair encoder aims to maximize the utility in a classification or decoding task. Previous work has mostly accomplished this in group fairness settings \cite{edwards2015censoring, madras2018learning}. For example, Edwards et al. \cite{edwards2015censoring} were among the first to propose learning a classifier that achieves demographic parity adverserially.
More recently, a stream of studies have incorporated individual fairness metrics in fair representation learning \cite{ruoss2020learning, lahoti2019operationalizing, feng2019learning}. 
In Section \ref{sec:evaluation}, we implement our framework based on LAFTR model proposed in \cite{madras2018learning}, which uses group fairness metric. 
However, our ``prior knowledge" approach 
can be easily applied to other existing fair representation learning algorithms that enforce individual fairness constraints.

\textbf{Interpretability.} Only few of the studies in fair representations have considered the interpretability of the representations. Recently, He et al. \cite{he2019learning} have proposed a geometric method that removes correlations between data and any number of protected attributes. They argue that the obtained features are interpretable and the data is \emph{debiased}. The main shortcoming of this approach, however, is that features that are correlated with the protected attributes can, in fact, be correlated with the label also. Therefore, debiasing those features would result in great utility loss. Other works that have considered interpretability in algorithmic fairness, such as Dwork et al. \cite{dwork2020abstracting}, seek interpretability for the purpose of allowing a human to discard unfair models. We, instead, seek interpretability of fair representations to guard against human biases (e.g. model assisted decision-making), and to allow the data user utilize the data for tasks beyond prediction.

\section{Algorithm}
\label{sec:algorithm}
\begin{figure}
    \centering
    \includegraphics[width=3in, height=2in]{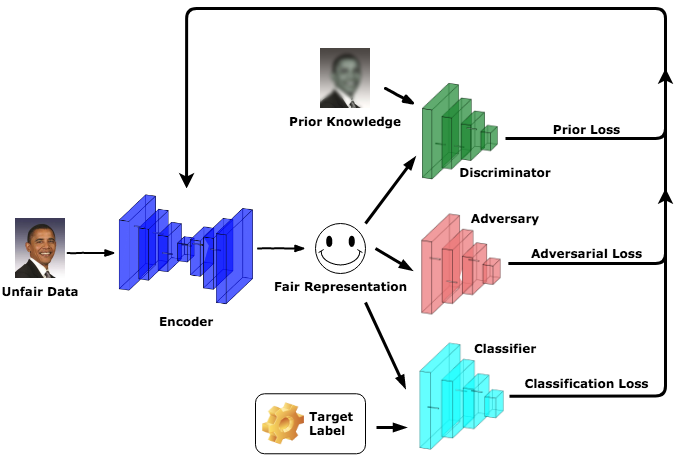}
    \caption{Overview of the proposed prior knowledge learning framework.}
    \label{fig:flowchart}
\end{figure}

An overview of our prior knowledge learning framework is presented in Figure \ref{fig:flowchart}. 
In this section, we present our method to learn interpretable fair representation in detail. 
We will use facial emotion recognition task as a running example for the illustration, where race is considered as a sensitive attribute.
Recent work shows that facial emotion recognition models can be easily biased towards certain groups, e.g. classifiers might tend to classify black faces as ``angry" \cite{garvie2016facial}. 
For a facial emotion recognition task, we hope the final fair representation can also be a face image instead of some random, meaningless image or 1-bit representation \cite{yang2019diversity}.

\begin{figure}[htbp]
\centering
\subfigure[]{
\begin{minipage}[t]{0.25\linewidth}
\centering
\includegraphics[width=0.8in]{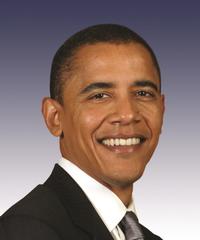}
\end{minipage}%
}%
\subfigure[]{
\begin{minipage}[t]{0.25\linewidth}
\centering
\includegraphics[width=0.8in]{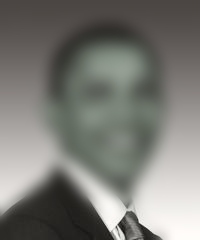}
\end{minipage}%
}%
\centering
\caption{Example of Processed Image}
\label{fig:blur} 
\end{figure}

We incorporate ``prior knowledge" into the existing state-of-art fair representation learning frameworks \cite{madras2018learning, ruoss2020learning, lahoti2019operationalizing, feng2019learning} to help learning interpretable fair representation. 
The key observation supporting our approach is that given an unfair representation, it is arguably easy for us to come up with another representation that ``we think” it is fair, especially for the image case.
For instance, for a face recognition task where race is considered a sensitive attribute, we can apply some image processing techniques, such as blurring, to the original images in order to remove the race information while maintain other human face features as much as possible. 
As as illustration, in Figure \ref{fig:blur}, most of people are not able to tell the race of the human in (b), but it is still recognizable that it's a man and he is smiling. 
We refer to these kind of blurred images as ``prior knowledge", as it represents the data owner's understanding of what an interpretable fair representation should look like. 
After we obtain these images that ``looks fair", we encourage the learned fair representations to be similar to these prior knowledge (but can have slight differences), so that the learned actual fair representation will be more realistic and recognizable instead of being some meaningless images. 
We emphasize that, although the blurred image in Figure \ref{fig:blur}(b) may still reveal some race information that can be recognized by an anthropologist or a carefully trained machine learning model, it still provides rich knowledge about how a fair face image \emph{might be structured}; a proper formulation of such prior knowledge will help regularize the 1-bit representation problem.

Specifically, we separate representation learning process into two stages: (1) \textit{Prior knowledge learning}, in which we train the encoder $f$ and a discriminator $\discriminator$ on prior knowledge set $\priorKnowledgeSet$ (e.g. some blurred images) to encourage the encoder to output an interpretable representation. (2) \textit{Enforcing fairness constraints}, in which we incorporate the trained encoder $f$ from the previous stage into the existing fair representation learning algorithms (\cite{madras2018learning, ruoss2020learning, lahoti2019operationalizing, feng2019learning}) to further enforce the different hard fairness constraints. 

In the first stage, we use canonical Wasserstein GAN \cite{arjovsky2017wasserstein} to facilitate the encoder learn the distribution underlying the prior knowledge set $\priorKnowledgeSet$:
$$
\min_f \max_\discriminator L_{Prior} (f, \discriminator) = E_{t \in T}[\discriminator(t)] - E_{x \in \originalSet}[\discriminator(f(x))]
$$
Besides, inspired by \cite{yang2019diversity}, we regularize prior knowledge learning by introducing a diversity loss term to increase the diversity of encoded images and address the notorious mode-collapse problem in Conditional GAN \cite{gauthier2014conditional}.
The diversity loss term can be written as 
$$
\max_{f}L_{div}(f) = E_{u, v \in \originalSet}
\left[ \frac{||f(u)-f(v)||}{||u-v||} \right]
$$
The proposed regularization term is mainly intended to ensure the prior knowledge learning itself does not fall into some trivial solution. 
Therefore, the full objective function for prior knowledge training is $\min_f \max_\discriminator L_{Prior}(f, \discriminator) - \lambda L_{div}(f)$. 

In the enforcing fairness constraints stage, we fine-tune the trained encoder $f$ from previous stage and enforce the fairness constraints.
Different target fairness constraints  will require different objective functions in this stage. 
In this paper, we take the LAFTR model proposed in \cite{madras2018learning} to enforce demographic parity for illustration as well as the experiments in Section \ref{sec:evaluation}, but the framework we proposed is general. 
Denote $A$, $g$, $h$, $L_{clf}$, $L_{adv}$ as the 
sensitive attribute, classifier and  adversary, classification loss and adversarial loss respectively in the original paper of LAFTR, the objective function for the second stage is:
$$
\min_{f, g, \discriminator}\max_{h} E_{X, Y, A}[L(f, g, h, \discriminator)]
$$
where 
\begin{align}
L(f, g, h, \discriminator) 
&= \alpha L_{clf}(g(f(X, A)), Y) \\
&+ \beta L_{Adv}(h(f(X, A)), A) \\
&+ \gamma L_{Prior}(f, \discriminator)
\end{align}
We fine-tune $f$ by solving the above optimization problem to enforce demographic parity, which is mainly ensured by $L_{adv}$. 
However, since the encoder $f$ is already trained to encode the original images into representations similar to $T$, the hope is that its parameter will not change too much after fine-tuning. 
The final encoder $f$ will output actual fair images that might be slightly different from $T$ (e.g. fine adjustment for face shape to hide subtle racial information), but there are high chances that the fair representations are realistic and recognizable if ``prior knowledge" is not too far from the true answer. 
We also penalize unrealistic images using the loss term from the first stage, $L_{prior}$, to ensure the learned representation does not deviate too much. 

\section{Evaluation}
\label{sec:evaluation}
\subsection{Experimental Setup}
\paragraph{Datasets.} We evaluate our methods with two synthetic datasets -- \texttt{ColorMNIST} and \texttt{Dsprite} \cite{dsprites17}. 
We split each dataset into two disjoint parts, where for the training sets, the labels have strong (but spurious) correlation with some specific ``sensitive" attribute, while for the test sets there are no such correlations. 
The details of constructing these two synthetic datasets will be discussed in Section \ref{sec:colormnist} and Section \ref{sec:dsprite},  respectively. 
\paragraph{Models. }We implement different neural networks as encoder, distriminator, adversary, and classifier. Some of the networks are adapted from the LAFTR \cite{madras2018learning} by adjusting the input/output shape of their layer to fit our tasks.
We use encoder that has 2 convolutional layers and 1 batch normalization layer for both datasets. 
\paragraph{Metrics.} To assess our framework and compare with previous methods, we train a 3-layer MLP on the original unfair dataset as well as fair representations outputted by encoders learned under either original LAFTR or our framework. 
We quantitatively assess the effectiveness of different algorithms by classifier performance and the demographic parity in accuracy. The demographic parity is defined as the largest difference of the classifier accuracies grouped by protected attributes. 
We also qualitatively measure the interpretability of fair representations by visual inspection. We leave developing quantitative metrics for measuring interpretability as future work.  

\subsection{Evaluation with ColorMNIST}
\label{sec:colormnist}
Different from the conventional MNIST dataset, our \texttt{ColorMNIST} has color on the digits, which analogous to the protected attributes in real-life image datasets such as race. Inspired by \cite{arjovsky2019invariant},
we first assign binary labels to all of the MNIST images based on whether the digit is larger than 4.5 (1) or less than 4.5 (0). This label is denoted as $\Tilde{y}$. We flip this label by a probability of 0.25 as label $y$. Finally, we obtain the color label $z$ by flipping $y$ for a probability of 0.2. 
In the training set, we color the images as red if $z$ is 1, and as green otherwise. 
In the test set, however, we randomly color the digits so there are no correlation between labels and colors. 
The prior knowledge we construct is a set of digit images where the shape of the digits sans-color, as that is how ``we think" what a fair representation should look like. 

\begin{figure}[htbp]
\centering
\subfigure[Biased \texttt{ColorMNIST}]{
\begin{minipage}[t]{0.25\linewidth}
\centering
\includegraphics[width=0.8in]{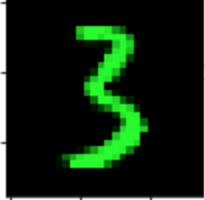}
\end{minipage}%
}%
\subfigure[LAFTR]{
\begin{minipage}[t]{0.25\linewidth}
\centering
\includegraphics[width=0.8in]{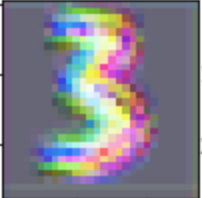}
\end{minipage}%
}%
\subfigure[PriorTraining (ours)]{
\begin{minipage}[t]{0.25\linewidth}
\centering
\includegraphics[width=0.8in]{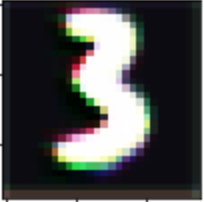}
\end{minipage}%
}
\centering
\caption{Example of \texttt{ColorMNIST} and encoded fair representations}
\label{fig:mnist} 
\end{figure}

We evaluate our framework by training an encoder through pipelines described in Section \ref{sec:algorithm} on \texttt{ColorMNIST}.
In Figure \ref{fig:mnist}, we observe that our fair representation most visually resembles the original MNIST dataset, which has no color and the shape of digit is much more clear than the representation outputted by original LAFTR. 
From Table \ref{tb:mnist}, we can  see that by training classifiers with representations incorporated with prior knowledge, we obtain slightly better model performance and higher fairness level than LAFTR. 
We note that the improvement of model performance and fairness level is mainly due to the high quality of our prior knowledge, and in the image domain it is relatively easy to obtain such good prior knowledge. 

\begin{table}[]
\centering
\begin{tabular}{|cccc|}
\hline
Algorithms                    & Training Acc & Test Acc & Demographic   Parity \\ \hline
Naïve                         & 0.8169       & 0.5831   & 0.539                \\ 
LAFTR                   & 0.7488       & 0.7622   & 0.026                \\ 
PriorTraining (ours)                 & 0.7862       & 0.7824   & 0.004                \\ \hline
Random   Guessing             & 0.5          & 0.5      & 0                    \\ 
Optimal Model   (theoretical) & 0.75         & 0.75     & 0                    \\ \hline
\end{tabular}
\caption{Model performance and fairness results for classifiers trained on representations provided by our proposed framework and original LAFTR, and naive training on unfair \texttt{ColorMNIST}.}
\label{tb:mnist}
\end{table}

\subsection{Evaluation with Dsprite}
\label{sec:dsprite}
\texttt{Dsprite} \cite{dsprites17} is a synthetic image dataset where each image has 6 unique descriptors. These 6 descriptors are color (while), shape (square, circle, heart), scale (6 equally spaced values $[0.5, 1]$),  orientation (40 equally spaced values $[0, 2\pi]$), x-position (32 equally spaced values $[0, 1]$) and y-position (32 equally spaced values $[0, 1]$). 
We define classifying the shape of the the images as the classification task, and consider the scale of the object as sensitive attribute.

\begin{figure}[htbp]
\centering
\subfigure[Shape is heart]{
\begin{minipage}[t]{0.3\linewidth}
\centering
\includegraphics[width=1.5in]{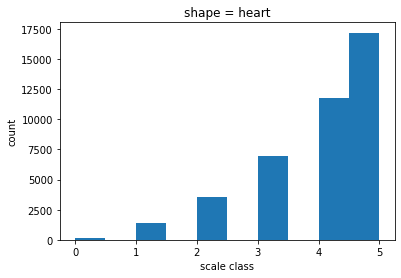}
\end{minipage}%
}%
\subfigure[Shape is circle]{
\begin{minipage}[t]{0.3\linewidth}
\centering
\includegraphics[width=1.5in]{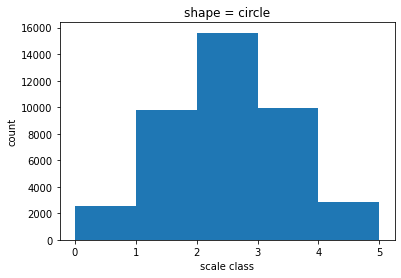}
\end{minipage}%
}%
\subfigure[Shape is square]{
\begin{minipage}[t]{0.3\linewidth}
\centering
\includegraphics[width=1.5in]{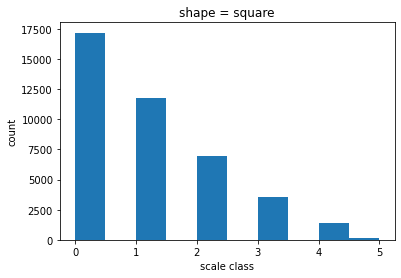}
\end{minipage}%
}
\centering
\caption{Distributions of scales based on shape.}
\label{fig:distribution} 
\end{figure}

We sample the training data in the following way to construct an artificial correlation from the dataset: if the shape is a heart, then the scale is assigned according to power law distribution. If circle, then the scale will be assigned according to normal distribution. If square, then the scale will be assigned according to a mirrored power law. 
$$
\begin{array}{ll}
Pr[scale| shape=heart] \propto scale^{\alpha}  \\
Pr[scale| shape=circle] \propto \mathcal{N}(\mu,\sigma^{2}) \\
Pr[scale| shape=square] \propto (6-scale)^{\alpha} \end{array}
$$
where $\alpha=3$, $\mu = 3.5$, $\sigma = 1$. The constants are picked such that the distributions makes individuals with lower scales still possible to have a shape of heart, and vise the versa. 
The distributions are also shown in Figure \ref{fig:distribution}. 

Similar to the case for \texttt{ColorMNIST}, we construct the test data of \texttt{Dsprite} s.t. there are no correlation between the scale and the shape. 
By doing so, we simulate a biased data that might be used real life where the scale is a continuous variable that correlates with positive outcome of the classifier. 
In this case, the representation that ``we think" is fair is when all of the samples have similar scale. Hence, we assign an average scale to all of the images as the ``imagined" fair representations.

\begin{figure}[htbp]
\centering
\subfigure[Biased \texttt{Dsprite}]{
\begin{minipage}[t]{0.25\linewidth}
\centering
\includegraphics[width=1in]{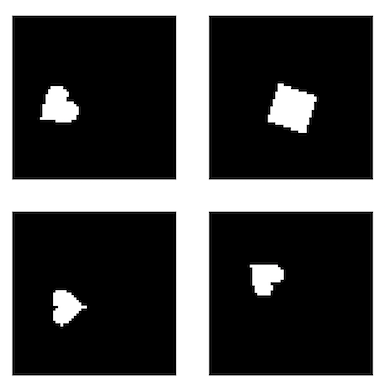}
\end{minipage}%
}%
\subfigure[LAFTR]{
\begin{minipage}[t]{0.25\linewidth}
\centering
\includegraphics[width=1in]{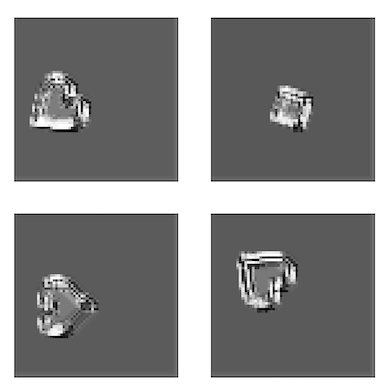}
\end{minipage}%
}%
\subfigure[PriorTraining (ours)]{
\begin{minipage}[t]{0.25\linewidth}
\centering
\includegraphics[width=1in]{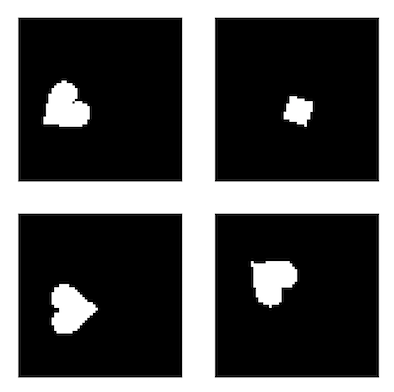}
\end{minipage}%
}
\centering
\caption{Example of \texttt{Dsprite} and encoded fair representations}
\label{fig:dsprite} 
\end{figure}

The biased \texttt{Dsprite} fair representations are shown in Figure \ref{fig:dsprite}. As we can see, the fair image encoded by our method is clearly more recognizable and closer to the original \texttt{Dsprite} than the one encoded by LAFTR. 
We also obtain slightly better training and test accuracy when we train a classifier on the representations encoded by our method, as shown in Table \ref{tb:dsprite}. 
The demographic parities are both very small in our method and LAFTR.
\begin{table}[]
\centering
\begin{tabular}{|cccc|}
\hline
Algorithms                    & Training Acc & Test Acc & Demographic   Parity \\ \hline
Naïve                         & 0.9874       & 0.858    & 0.227                \\ 
LAFTR                   & 0.9717       & 0.9525   & 0.01               \\ 
PriorTraining (ours)                & 0.9864       & 0.982    & 0.01                \\ \hline
Random   Guessing             & 0.33         & 0.33     & 0                    \\ 
Optimal Model   (theoretical) & 1            & 1        & 0                    \\ \hline
\end{tabular}
\caption{Model performance and fairness results for classifiers trained on representations provided by our proposed framework and original LAFTR, and naive training on unfair \texttt{Dsprite}.}
\label{tb:dsprite}
\end{table}

\section{Discussion \& Future Work}

We present a general framework for learning fair, yet interpretable representations. We experiment with two synthetic datasets, and show that our algorithm is superior compared to state-of-the-art baselines in terms of accuracy, fairness and interpretability.

In this work, we demonstrated our approach only with synthetic image data. However, tabular data is usually much more common in fairness-sensitive applications. Learning interpretable representation on tabular data will be an interesting future work. Currently, our framework does not directly apply to tabular data since it is typically more difficult to think of how a fair representation might ``look" like, especially when the dimension of the data becomes large and multiple columns are correlated with the sensitive attributes. 

Further, in this work, we only qualitatively measure the interpretability of the representations through visual inspection; it is important to come up with formal quantitative metrics for measuring interpretability. Because of the subjective nature of the concept of interpretability, we aim to survey human subjects to evaluate the representations in terms of interpretability; we will conduct survey experiments on platforms such as Amazon MTurk (AMT) to obtain quantitive scores for representation's interpretability. 

\newpage

\bibliographystyle{ieeetr}
\bibliography{ref}

\end{document}